\documentclass[10pt, a4paper]{article}

\usepackage[]{lrec-coling2024} 

\usepackage{times}
\usepackage{CJKutf8}
\usepackage{soul}
\usepackage{url}
\usepackage[utf8]{inputenc}
\usepackage[small]{caption}
\usepackage{graphicx}
\usepackage{xcolor}
\usepackage{amsmath}

\usepackage{amsthm}
\usepackage{booktabs}
\usepackage{algorithm}
\usepackage{latexsym}
\usepackage{graphicx} 
\usepackage{amsmath}
\usepackage{subfigure}
\usepackage{microtype}
\usepackage{algorithmic}
\usepackage[switch]{lineno}
\usepackage[most]{tcolorbox}
\usepackage{alltt}

\definecolor{forestgreen}{rgb}{0.13, 0.55, 0.13}

\tcbset{
    aibox/.style={
        colback=white, 
        colframe=blue!75!black, 
        colbacktitle=blue!85!black, 
        coltitle=white, 
        fonttitle=\bfseries,
        enhanced,  
        drop shadow=black!50!white,  
        boxrule=1pt,  
        boxsep=10pt,  
        left=10pt,  
        right=10pt,  
        top=6pt,  
        bottom=6pt,  
        title code={\path[tcb fill frame] ([xshift=-10pt]frame.west) -- (frame.north west) -- (frame.north east) -- ([xshift=10pt]frame.east) -- cycle;},  
        attach boxed title to top left={xshift=10pt, yshift*=-\tcboxedtitleheight/2},
        boxed title style={boxrule=0pt, frame code={}}  
    }
}

\newtcolorbox{AIbox}[2][]{aibox, title=#2, #1}

\title{A Paradigm Shift: The Future of Machine Translation Lies with Large Language Models}

\name{Chenyang Lyu\textsuperscript{1$\dagger$}, Zefeng Du\textsuperscript{2$\dagger$}, Jitao Xu\textsuperscript{3$\dagger$}\thanks{$\dagger$ Equal contribution.},
Yitao Duan\textsuperscript{3}, Minghao Wu\textsuperscript{4}, \\ 
{\bf \large Teresa Lynn\textsuperscript{1}, Alham Fikri Aji\textsuperscript{1}, Derek F. Wong\textsuperscript{2}, Siyou Liu\textsuperscript{2}, Longyue Wang\textsuperscript{5}}} 

\address{1. MBZUAI\\
2. University of Macau \\
3. NetEase Youdao \\
4. Monash University \\
5. Tencent AI Lab \\
chenyang.lyu@mbzuai.ac.ae, mc05583@umac.mo, xujt01@rd.netease.com, duan@rd.netease.com, \\vinnlywang@tencent.com\\}

\abstract{
Machine Translation (MT) has greatly advanced over the years due to the developments in deep neural networks. However, the emergence of Large Language Models (LLMs) like GPT-4 and ChatGPT is introducing a new phase in the MT domain. In this context, we believe that the future of MT is intricately tied to the capabilities of LLMs. These models not only offer vast linguistic understandings but also bring innovative methodologies, such as prompt-based techniques, that have the potential to further elevate MT. In this paper, we provide an overview of the significant enhancements in MT that are influenced by LLMs and advocate for their pivotal role in upcoming MT research and implementations. We highlight several new MT directions, emphasizing the benefits of LLMs in scenarios such as Long-Document Translation, Stylized Translation, and Interactive Translation. Additionally, we address the important concern of privacy in LLM-driven MT and suggest essential privacy-preserving strategies. By showcasing practical instances, we aim to demonstrate the advantages that LLMs offer, particularly in tasks like translating extended documents. We conclude by emphasizing the critical role of LLMs in guiding the future evolution of MT and offer a roadmap for future exploration in the sector.
 \\ \newline  }

\begin{document}

\maketitleabstract

\section{Introduction}

\begin{figure*}[!ht]
    \centering
    \includegraphics[width=0.8\linewidth]{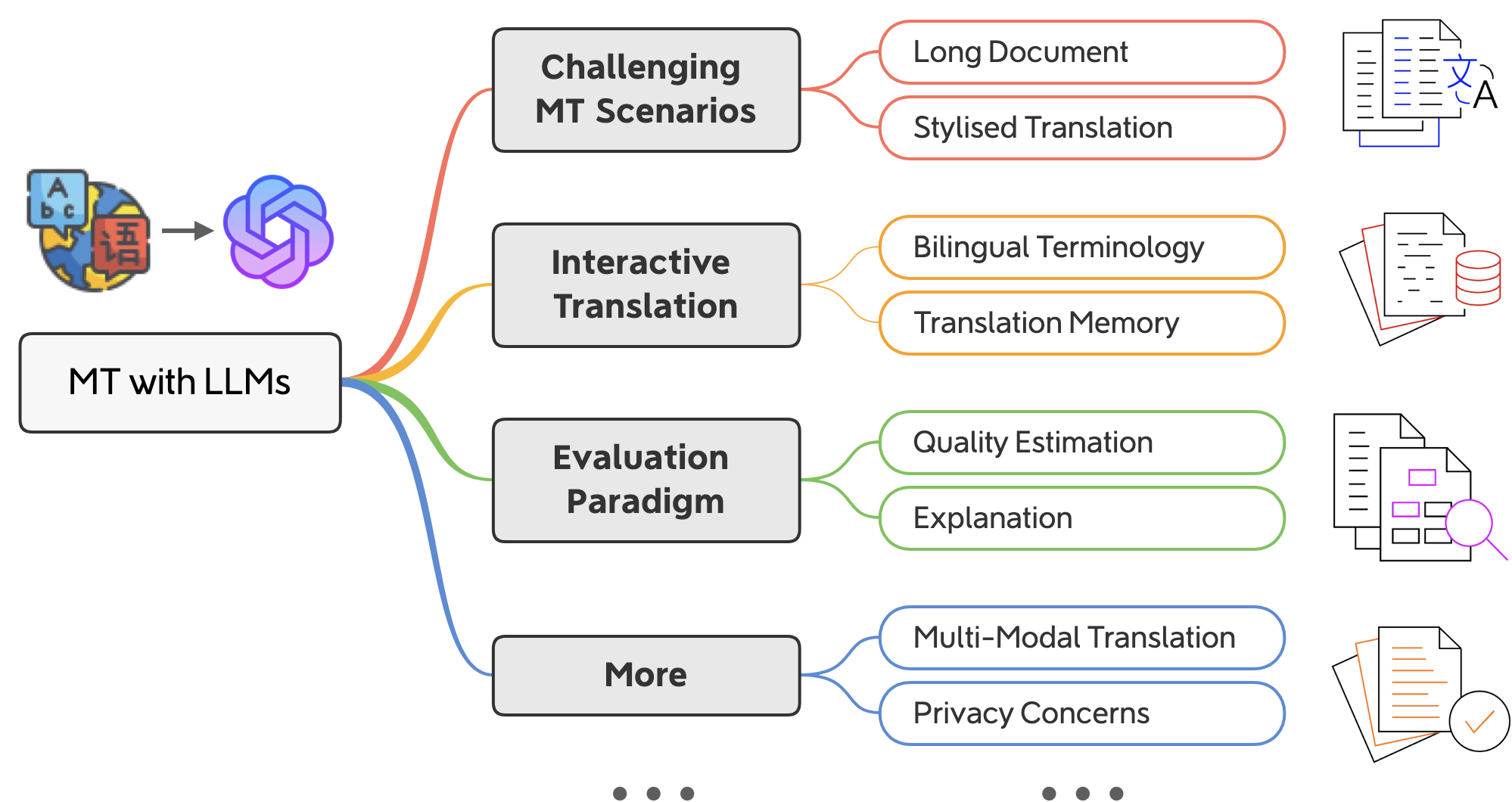}
    \caption{Interesting directions for MT using LLMs (e.g. GPT models), including challenging MT scenarios, interactive MT, new evaluation paradigm for MT using LLMs, etc.}
    \label{fig:directions}
\end{figure*}

\begin{figure}[!ht]
\centering
\begin{CJK}{UTF8}{gbsn}
\scalebox{0.9}{
\begin{AIbox}{GPT-4}
{\bf Prompt:} \\
\includegraphics[width=1\linewidth]{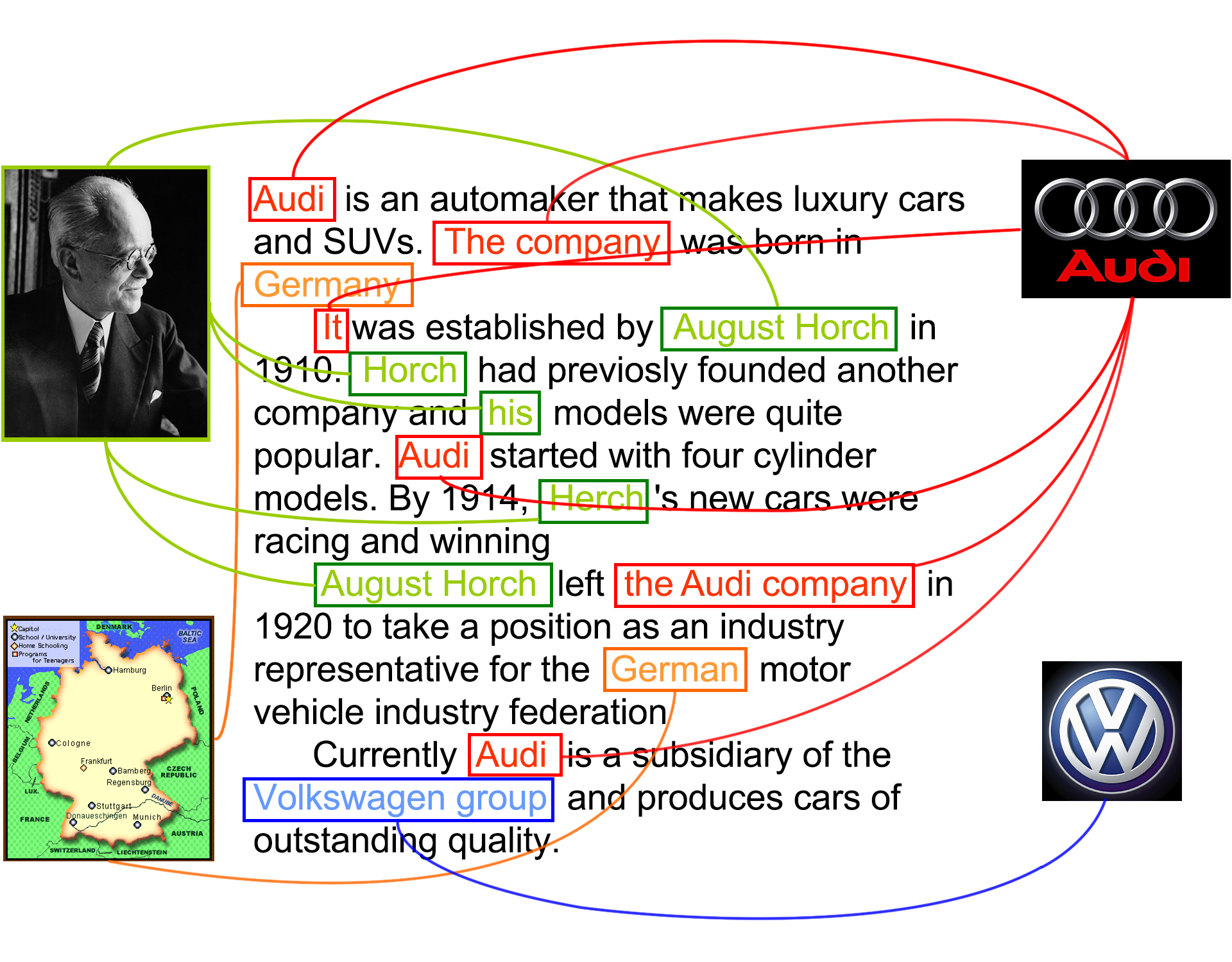}

{\bf Output:} \\
\small 奥迪是一家生产豪华汽车和SUV的汽车制造商。该公司起源于德国。\\
它由奥古斯特·霍希 (August Horch) 于1910年创立。霍希此前曾创立过另一家公司，其车型相当受欢迎。奥迪起初生产四缸车型。
到1914年，霍希的新车开始参加比赛并获胜。\\
奥古斯特·霍希于1920年离开奥迪公司，担任德国机动车行业联合会的行业代表。\\
目前，奥迪是大众集团的子公司，生产质量卓越的汽车。
\hfill
\end{AIbox}
}
\end{CJK}
\caption{An example of translating a document-level text from English to Chinese using GPT-4. We highlight the discourse phenomena using figures and lines, which are invisible to GPT-4.}
\label{fig:document-mt}
\end{figure}

Machine Translation~(MT) is a fundamental task in Natural Language Processing (NLP) that aims to automatically translate texts from one language to another~\cite{tsujii1986future_MT,DBLP:conf/coling/SatoN90_MT}. The performance and quality of MT systems have been significantly advanced from Statistical Machine Translation~(SMT)~\cite{zens2002phrase_smt,koehn2007moses} to Neural Machine Translation~(NMT)~\cite{cho2014learning_neural_smt,Bahdanau15neural} with the employment of machine learning techniques~\cite{Vaswani17attention,castilho2017neural,stahlberg2020neural,kocmi-etal-2022-findings}. Despite these advancements achieved within decades of research, MT still faces many challenges, such as dealing with idiomatic expressions, low-resource translation, handling rare words, and maintaining coherence and fluency in the translation~\cite{koehn2017six,wang2019discourse,yang2020survey-mt,haddow-etal-2022-survey-mt}.

Recently, the emergence of Large Language Models~(LLMs), such as GPT-3 and GPT-4~\cite{brown2020language-gpt3,chen2021evaluating-codex-gpt3,ouyangtraining-instructGPT,weifinetuned,hadi2023large}, has substantially reshaped the paradigm for MT from multiple dimensions. The zero-shot performance of LLMs on translation is even on par with strong fully supervised MT systems~\cite{jiao2023chatgpt,robinson2023chatgpt_mt,moslem2023adaptive,pang2024salute}. More importantly, LLMs can also be used in various scenarios beyond MT such as question answering and style transfer~\cite{bang2023multitask_eval_chatgpt,laskar-etal-2023-systematic_eval_chatgpt,li2023bactrian}, which enables novel scenarios and provide rooms for exploration for MT. Besides these opportunities, LLMs-based MT poses new challenges such as privacy-related issues that require new directions and methodologies to be addressed. 

In this paper, with the aim of extending the scope of MT with the incorporation of the superior capability of LLMs, we present discussions on several potentially interesting and promising novel directions for LLMs-based MT, including challenging translation scenarios such as long-document translation and stylised translation, interactive translation, and Translation Memory~(TM) based MT, potential new evaluation paradigms of translation quality using LLMs, as well as privacy concerns for LLMs-based MT and some other interesting directions.

Among the aforementioned directions, challenging translation scenarios include long-document translation~\cite{maruf2021survey_dmt} which requires the translation of long documents with thousands of words or even longer and stylised translation that aims to preserve the stylistic features of the source text or inject specific language styles in the translation output, such as the tone, register, formality, genre, etc~\cite{Sennrich16controlling,Niu20controlling}. Interactive translation~\cite{Knowles16interactive,Santy19inmt} aims to facilitate the collaboration and feedback between human translators and MT systems, such as through chatbots or question-answering systems. TM-based MT~\cite{Bulte19fuzzy,Xu20boosting} tends to make use of similar or relevant translations to improve the quality of the translation output, which is increasingly important in the era of LLMs with the use of in-context learning that can inform LLMs the necessary knowledge in the demonstration examples~\cite{moslem2023adaptive}. Moreover, we explore multi-modal translation with LLMs which translates source texts with images as extra context~\cite{sulubacak2020multimodal,yao-wan-2020-multimodal}. Furthermore, we discuss the potential new evaluation paradigm for LLMs-based MT that aims for a more accurate and efficient evaluation of MT systems from various aspects instead of only evaluating the similarity between system outputs and references~\cite{kocmi2023large,liu2023gpteval}. Besides the new directions and methodologies, we also discuss the privacy concerns in MT using LLMs and propose basic privacy-preserving methods to mitigate the risks. Privacy in NLP and LLMs is becoming increasingly important~\cite{klymenko-etal-2022-differential,privatenlp-2022-privacy-nlp,li2023privacypreserving}, as LLMs may inadvertently reveal sensitive information in the source text or the translation output when using LLMs for translation. In addition, we present discussions on some other new scenarios for LLMs-based MT such as personalized MT and low-resource language translation using LLMs.

To preliminarily investigate the feasibility of the interesting directions mentioned above, we present some empirical evidence in Section~\ref{sec:results} and examples using various LLMs such as LLaMA and GPT-4 for MT under various scenarios, demonstrating the feasibility of the directions. This position paper demonstrates the potentials of the prospective new directions and methodologies for enhancing the quality and diversity of MT output, as well as the importance and challenges of privacy-preserving in MT using LLMs. We conclude by highlighting the opportunities and challenges for future research in MT using LLMs and suggesting potential directions for further exploration.




\section{Challenging MT Scenarios}
\subsection{Long-Document Translation}

The majority of MT applications have traditionally concentrated on sentence-level translation~\cite{post2023escaping}, which can sometimes lead to translations that are devoid of context and coherence when translating long-documents that might contain thousands of words with complex structure. Recent years have seen a growing interest in document-level translation, a task of critical importance that involves the translation of entire documents, but also presents unique challenges~\cite{wang-etal-2017-exploiting-cross,voita-etal-2019-good,wang2018translating,zhang2022multilingual,jiang-etal-2022-blonde,wu-etal-2023-document,wang2023document,wu2024importance}. Surprisingly, LLMs have shown potentials in modeling exceptionally long texts with complex discourse structures, suggesting that they could be instrumental in advancing the field of document-level translation~\cite{wang2023document,he2023exploring_maps_mt,wu2024adapting}. Figure~\ref{fig:document-mt} illustrates an example of GPT-4 translating a document, where the discourse phenomena such as pronouns must be well translated in order to maintain the document structure. Furthermore, with the introduction of larger context window for LLMs~\cite{tworkowski2023focused-long-llama,xiong2023effective-long-llama}, the translation of longer documents such as novels and books with LLMs become more and more feasible.

\subsection{Stylised Translation}

Stylised Translation refers to the ability of generating translations that match a specific style or genre~\cite{wang2022controlling}, such as formal or informal expression~\cite{Sennrich16controlling}, poetry or prose, different dialects or registers, etc. This can be achieved by training MT systems on multi-parallel data that contain translations in different styles or genres, or by using style transfer techniques~\cite{yang2018unsupervised} that can transform a given translation into a desired style. Stylised Translation has many potential applications, such as in marketing, literature, or cultural preservation.

However, Stylised Translation is difficult to achieve before the presence of LLMs as there lacks such multi-parallel corpora for Stylised Translation to fit various styles while the zero-shot ability of LLMs makes these tasks seamlessly achievable.  It can be achieved by directly prompting LLMs to translate the text with a specific style expressed by natural language or firstly let LLMs translate the original text and then stylise the translation output. We present an example of translating an introduction for the Olympic Games from Wikipedia from English to Chinese while following a poetic style in Figure~\ref{fig:example_Stylised_translation}. This example shows GPT-4 can handle translation with a poetic style while keeping the semantic information of the original text, which can be hardly achieved by traditional MT systems. 

Nevertheless, Stylised Translation also presents a variety of obstacles. Among these challenges, one notable issue is to determine the best approach to systematically define and quantify various styles or genres on a larger scale. Another challenge is how to evaluate the quality of Stylised Translation, as traditional evaluation metrics may not be sufficient to capture the diversity of stylistic variations. Overcoming these challenges requires interdisciplinary collaboration between linguists, literary scholars, and computer scientists.

\begin{figure}[!ht]
\centering
\begin{CJK}{UTF8}{gbsn}
\scalebox{0.9}{
\begin{AIbox}{GPT-4}
{\bf Prompt:} 
\small Please translated the text below to Chinese following poetic style: "The Ancient Olympic Games were a religious and athletic festival that took place every four years at the sanctuary of Zeus in Olympia, Greece. Participants came from various city-states and kingdoms of Ancient Greece, competing in mainly athletic sports, but also combat sports such as wrestling and the pankration, as well as horse and chariot racing events. It is believed that during the Games, all conflicts among participating city-states were suspended until the end of the festival, which was known as the Olympic peace or truce. However, this idea is considered a modern myth as the Greeks never actually suspended their wars. Nevertheless, the truce did provide safe passage for religious pilgrims traveling to Olympia, who were protected by Zeus."\\

{\bf Output:} \\
\small 古奥林匹克运动会，四年一度，希腊奥林匹亚，\\
宙斯圣地，宗教盛典，竞技同场。\\
诸城邦国，齐聚一堂，力争上游，\\
田径角逐，搏击角力，马车竞速。\\
奥运和平，节庆期间，战争暂停，\\
虽为现代，神话传说，实则不然。\\
宙斯庇佑，朝圣者行，安全抵达，\\
奥林匹克，和平永驻，共祈祝愿。\\
\hfill
\end{AIbox}
}
\end{CJK}
\caption{An example of prompting GPT-4 to translate texts from English to Chinese using poetic style. GPT-4 can generate translations in a poetic style, and its format is also more in line with poetry, while the semantic information of the original English text is also preserved in the output translation.}	
\label{fig:example_Stylised_translation}
\end{figure}

\section{Interactive Translation}

\begin{figure}[!ht]
\centering
\begin{CJK}{UTF8}{gbsn}
\scalebox{0.9}{
\begin{AIbox}{GPT-4}
\small{\bf Prompt:} Please translate the text below to Chinese: "Jordan played college basketball for three seasons under coach Dean Smith with the North Carolina Tar Heels. As a freshman, he was a member of the Tar Heels' national championship team in 1982. Jordan joined the Bulls in 1984 as the third overall draft pick, and quickly emerged as a league star, entertaining crowds with his prolific scoring while gaining a reputation as one of the game's best defensive players. His leaping ability, demonstrated by performing slam dunks from the freethrow line in Slam Dunk Contests, earned him the nicknames "Air Jordan" and "His Airness"."\\

{\bf Output:} 乔丹在北卡罗来纳州\textcolor{red}{塔尔希尔}队的教练\textcolor{red}{迪恩·史密斯}指导下，为期三个赛季的大学篮球生涯。作为一名新生，他在1982年成为了\textcolor{red}{塔尔希尔}队全国冠军团队的一员。1984年，乔丹作为第三顺位选秀被\textcolor{red}{公牛}队选中，并迅速成为联盟明星，以高效得分娱乐观众，同时也被认为是比赛中最好的防守球员之一。他的跳跃能力，通过在扣篮大赛中从罚球线起跳完成扣篮而展示出来，为他赢得了“\textcolor{red}{飞人乔丹}”和“\textcolor{red}{空中魔术师}”的绰号。\\
\hfill
\tcbline
{\bf Prompt}:Please translate it to Chinese while keep the people's name (such as Jordan) and organisation's name (such as Chicago Bull) in English.\\

{\bf Output}:乔丹在北卡罗来纳州\textcolor{blue}{Tar Heels}队的教练\textcolor{blue}{Dean Smith}指导下，为期三个赛季的大学篮球生涯。作为一名新生，他在1982年成为了\textcolor{blue}{Tar Heels}队全国冠军团队的一员。1984年，乔丹作为第三顺位选秀被\textcolor{blue}{Chicago Bulls}队选中，并迅速成为联盟明星，以高效得分娱乐观众，同时也被认为是比赛中最好的防守球员之一。他的跳跃能力，通过在扣篮大赛中从罚球线起跳完成扣篮而展示出来，为他赢得了“\textcolor{blue}{Air Jordan}”和“\textcolor{blue}{His Airness}”的绰号。\\

\end{AIbox}
}
\end{CJK}
\caption{An example of using GPT-4 in an interactive way for MT. GPT-4 can adjust the translation results according to the user's input. Named entities (in \textcolor{red}{red}) ajusted by GPT-4 are marked in \textcolor{blue}{blue}.}	
\label{fig:example_interactive_mt}
\end{figure}

\begin{figure}[!ht]
\centering
\begin{CJK}{UTF8}{gbsn}
\scalebox{0.9}{
\begin{AIbox}{GPT-4}
\small{\bf Prompt:} Please translate the following sentence into French. Here is a similar translation you can use:

En cas d'atteinte sévère ou si le patient est dans l'incapacité d'avaler les comprimés (par ex. patients alimentés par sonde), il est recommandé de débuter le traitement par une administration intraveineuse de ciprofloxacine jusqu'à ce qu'un relais par voie orale soit possible. \\

In severe cases or if the patient is unable to take oral suspension (e.g- patients on enteral nutrition), it is recommended to commence therapy with intravenous ciprofloxacin until a switch to oral administration is possible.\\

{\bf Output \textit{without TM}:} Dans les cas graves ou si le patient \textcolor{red}{ne peut pas prendre de} suspension orale (par \textcolor{red}{exemple,} les patients sous nutrition entérale), il est recommandé de commencer le traitement par \textcolor{red}{ciprofloxacine intraveineuse} jusqu'à ce qu'un passage à l'administration orale soit possible.\\

{\bf Output:} Dans les cas graves ou si le patient \textcolor{blue}{est incapable de prendre une} suspension orale (par \textcolor{blue}{ex.} les patients en nutrition entérale), il est recommandé de commencer le traitement par \textcolor{blue}{une administration intraveineuse de ciprofloxacine} jusqu'à ce qu'un passage à l'administration orale soit possible.
\hfill
\end{AIbox}
}
\end{CJK}
\caption{An example of prompting GPT-4 to translate text from
English to French using a similar translation retrieved from TM. Benefits directly taken by GPT-4 from the given similar translation is marked in \textcolor{blue}{blue}, in contrast to the translation in \textcolor{red}{red} without using TMs.}	
\label{fig:example_tm-based_mt}
\end{figure}

\begin{figure}[]
\centering
\begin{CJK}{UTF8}{gbsn}
\scalebox{0.8}{
\begin{AIbox}{GPT-4-Vision}
{\bf Prompt:} \\

\includegraphics[width=1\linewidth]{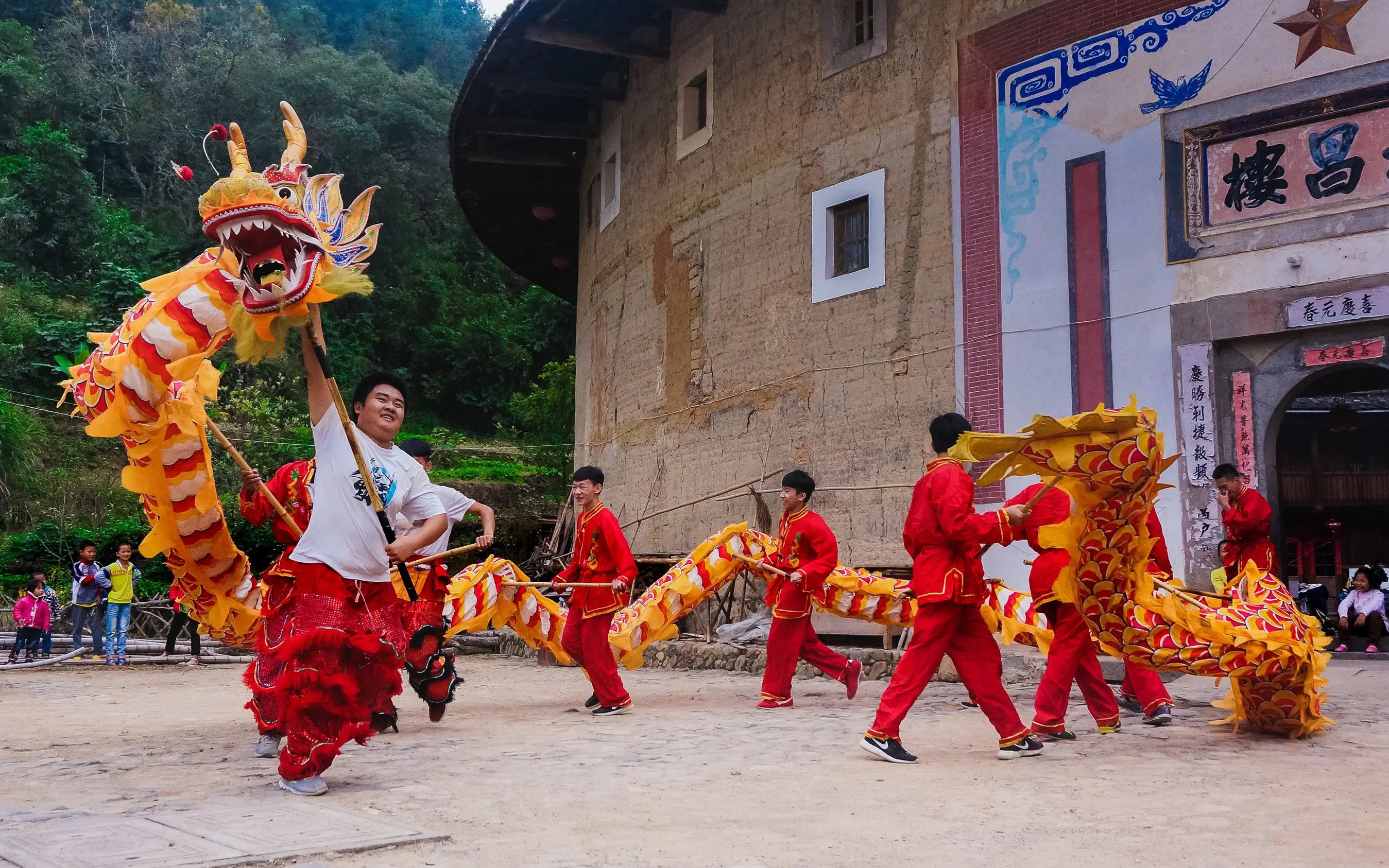}

Please translated the text below to English while considering the context of the image: "欢乐的春节舞龙巡游，充满了喜庆和活力" \\

{\bf Output:} 
\small The text translates to: \\ 
"Joyful Spring Festival loong dance parade, full of festivity and vitality." \\
\hfill
\end{AIbox}
}
\end{CJK}
\caption{An example of propmpting GPT-4-Vision for translating a sentence in Chinese to English while considering an associated image which shows a scene of celebrating new year.}
\label{fig:multimodal-mt}
\end{figure}

\begin{figure}[]
\centering
\begin{CJK}{UTF8}{gbsn}
\scalebox{0.8}{
\begin{AIbox}{GPT-4-Vision}
{\bf Prompt:} \\

\includegraphics[width=\linewidth]{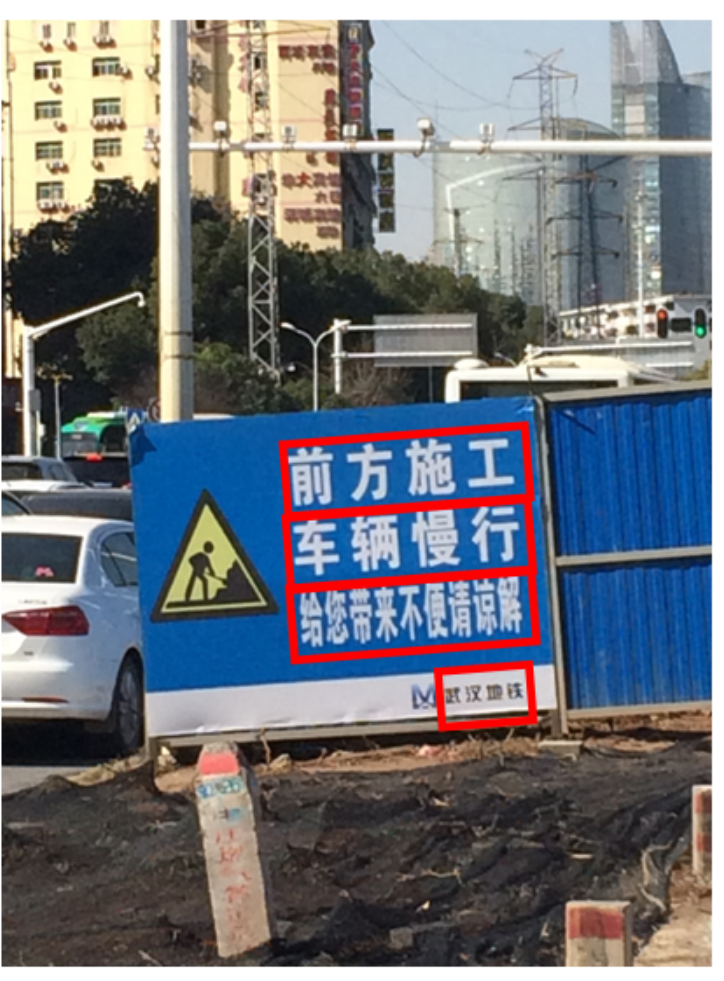}

Please translate the text in the image that I saw then driving a car into English:
 \\

{\bf Output:} 
\small The text in the image contains Chinese characters. Here are the translations: \\

"前方施工" translates to "Construction ahead." \\
"车辆绕行" translates to "Vehicles detour." \\
"禁止停车" translates to "No parking." \\
"土石工程" translates to "Earthwork project." \\
\hfill
\end{AIbox}
}
\end{CJK}
\caption{An example of a Chinese roadside construction sign translated by GPT-4-Vision: highlighting warnings for upcoming construction, a detour for vehicles, and a no-parking zone.}
\label{fig:multimodal-mt-1}
\end{figure}

\begin{figure}[]
\centering
\begin{CJK}{UTF8}{gbsn}
\scalebox{0.86}{
\begin{AIbox}{GPT-4}
\small{\bf Prompt:} Please translate this tweet into Chinese: "For the next 12 months, Twitter will keep none of the money.\\
You will receive whatever money we receive, so that'\$ 70\% for subscriptions on ios \& Android (they charge 30\%) and -92\% on the web (could be better, depending on payment processor）\\
After the first year. ios \& Android fees drop to 15\% and we will add a small amount on top of that, depending on volume.\\
We will also help promote your work. Our goal is to maximize creator prosperity.\\
At any point, you can leave our platform and take your work with you. Easy in, easy out."\\

{\bf Output:} 在接下来的12个月里，Twitter不会保留任何钱。\\
我们收到的钱你们将全部获得，所以在iOS和Android上的订阅收入是70\%（它们收取30\%的费用），在网页上是92\%（取决于支付处理器，可能会更好）\\
第一年过后，iOS和Android的费用降低到15\%，我们会在此基础上增加一小部分，具体取决于交易量。\\
我们还将帮助推广你的作品。我们的目标是最大化创作者的繁荣。\\
在任何时候，你都可以离开我们的平台并带走你的作品。轻松进入，轻松退出。\\
\tcbline
{\bf Prompt:} How good is the translation above with some specific aspects you can think of?\\

{\bf Output:} The translation above is generally accurate and conveys the main ideas of the original tweet. However, there are a few aspects that could be improved:

1. Consistency in terms: In the original tweet, "money" is used consistently, while in the translation, both "钱" and "收入" are used. It would be better to stick with one term, such as "收入" for consistency.

2. Clarity: The part "-92\% on web" in the original tweet might be a typo and should be "92\% on web." The translation follows the original text, but it could cause confusion for readers. It would be better to clarify this point in the translation.

3. Punctuation: The original tweet uses a mix of punctuation marks, such as "\$" and "\%." In the translation, it would be better to use standard Chinese punctuation marks, like "\%" for percentages.

4. Formatting: The translation is presented as a single paragraph, while the original tweet has multiple lines and breaks. To improve readability, the translation could be formatted similarly to the original tweet.

Overall, the translation is good but could benefit from some minor adjustments to improve consistency, clarity, punctuation, and formatting.
\hfill
\end{AIbox}
}
\end{CJK}
\caption{An example of using GPT-4 for evaluating translation output. We let GPT-4 generate a translation text, and then ask it to introduce how to evaluate a translation text. GPT-4 provides standards and evaluates the translation text accordingly. }	
\label{fig:overall_figure_eval}
\end{figure}

\begin{figure}[]
\centering
\begin{CJK}{UTF8}{gbsn}
\scalebox{0.86}{
\begin{AIbox}{GPT-4}
\small{\bf Prompt:} Please translate this text into Chinese: "John Smith, the CEO of Boeing Corporation, has offered to pay \$10 million for the acquisition of Alphabet Corporation. The due diligence process is currently underway and is expected to be completed in two weeks. Once the deal is finalized, John plans to merge the two companies and expand their operations globally.\\

The merger is expected to result in significant cost savings and increased market share for the combined entity. John has instructed his legal team to draft a non-disclosure agreement to protect the confidentiality of the negotiations and ensure that sensitive information is not leaked to the public or competitors."\\

{\bf Output:} \textcolor{red}{约翰·史密斯（John Smith）}，\textcolor{red}{波音公司（Boeing Corporation）}的首席执行官，已提议支付\textcolor{red}{1000万}美元收购\textcolor{red}{Alphabet Corporation}。尽职调查过程目前正在进行中，预计将在两周内完成。一旦交易敲定，\textcolor{red}{约翰}计划将两家公司合并，并将其业务扩展到全球。\\

这次合并预计将为合并后的实体带来显著的成本节约和市场份额的增长。\textcolor{red}{约翰}已指示他的法律团队起草一份保密协议，以保护谈判的机密性，并确保敏感信息不会泄露给公众或竞争对手。\\
\tcbline
{\bf Prompt:} Please translate this text into Chinese: "\#P1\#, the CEO of \#ORG1\#, has offered to pay \$\#NUM1\# million for the acquisition of \#ORG2\#. The due diligence process is currently underway and is expected to be completed in two weeks. Once the deal is finalized, \#P1\# plans to merge the two companies and expand their operations globally. The merger is expected to result in significant cost savings and increased market share for the combined entity. \#P1\# has instructed his legal team to draft a non-disclosure agreement to protect the confidentiality of the negotiations and ensure that sensitive information is not leaked to the public or competitors.\\

{\bf Output:} \textcolor{blue}{\#P1\#}，\textcolor{blue}{\#ORG1\#}的首席执行官，已提议支付\textcolor{blue}{\#NUM1\#百万}美元收购\textcolor{blue}{\#ORG2\#}。尽职调查过程目前正在进行中，预计将在两周内完成。一旦交易达成，\textcolor{blue}{\#P1\#}计划将两家公司合并并将其业务扩展至全球。预计合并将为合并后的实体带来显著的成本节约和市场份额增长。\textcolor{blue}{\#P1\#}已指示其法律团队起草一份保密协议，以保护谈判的保密性，确保敏感信息不会泄露给公众或竞争对手。\\
\hfill
\end{AIbox}
}
\end{CJK}
\caption{An example of privacy issue using GPT-4 for MT. The above one is the example where the input is not anonymized, thus containing name information, business data, etc (in \textcolor{red}{red}). The bottom one is the example where the sensitive information in the input is anonymized (in \textcolor{blue}{blue}). }	
\label{fig:overall_figure_privacy}
\end{figure}

\subsection{Bilingual Terminology}

Interactive Translation~\cite{Santy19inmt,jiao2023parrot} allows users to actively participate in the translation process, either by correcting or refining automatic translations or by providing feedback on the translation quality. This can be achieved by integrating MT systems based on LLMs that has superior conversational capability for language understanding and generation with interactive user interfaces, such as chatbots, that allow users to engage with the translation process in real-time to provide feedback and more specific requirements such as specific translations of certain terminologies. Interactive Translation can help to improve the accuracy and fluency of the translations, especially in cases where the source language is ambiguous or the domain knowledge is limited.

However, interactive MT also raises several challenges. One challenge is how to design user interfaces that are intuitive, user-friendly, informative and flexible. Another challenge is how to incorporate user feedback into the translation process in a principled and effective way. Overcoming these challenges requires insights from human-computer interaction, NLP, and user experience design. Figure~\ref{fig:example_interactive_mt} illustrates an example of prompting GPT-4 in an interactive way for MT, where the specific translation requirement of named entities is provided.


Previous studies on conventional TM-based MT has also shown that conventional Transformer-based NMT system already shows the ability to make use of new TMs that the model has never seen during training to largely improve domain-specific translation during inference~\cite{Xu20boosting,Xu22bilingual}. This indicates that conventional NMT systems learn to understand the relationship between a given source sentence and a similar translation and to select useful information from the given similar translation, rather than remember sentences seen during training. This ability is, to some extent, similar to the ICL ability of LLMs. However, to the best of our knowledge, there does not exist research works focusing on finding the relationships between these two abilities.

\subsection{Translation Memory-based MT}

TM has been used for decades to help human translators in basic Computer-Aided Translation systems. The general process of using TM in MT is, for a sentence to be translated, to first search for similar translations in TM using, for instance, fuzzy matching techniques, then revised or edit the retrieved similar translation in order to obtain a high-quality translation. TM-based MT has already been integrated into conventional NMT systems~\cite{Bulte19fuzzy,Xu20boosting,Cai21neural}. Since LLMs have emerged with the In-Context Learning (ICL) ability that they can learn specific tasks through task examples given in the prompt. The use of retrieved similar or relevant sentence pairs~\cite{Pham20priming} seems to be a natural fit to few-shot prompting techniques when performing MT using LLMs~\cite{Vilar22prompting,moslem2023adaptive}.

However, existing works so far have mostly used randomly selected translation examples as prompts and suggest that using semantically similar examples does not significantly further improve the translation performance \cite{Vilar22prompting,Zhu23multilingual}. Most of these works used sentence-level embedding built by an external language model to retrieve similar examples via an embedding similarity search. On the contrary, other studies using lexical fuzzy matches to retrieve similar translations have shown significant improvements~\cite{moslem2023adaptive}. Therefore, the conclusion about the effectiveness of using similar translations in MT using LLMs still remains unclear. Since TMs can provide useful domain and style information that can directly help LLMs to generate translations that better meet the translation requirement, it is a promising direction to further study how to better integrate TMs into LLMs for MT. Figure~\ref{fig:example_tm-based_mt} illustrates an example of prompting LLM both without and with TMs where using TMs directly improves the translation quality.


\section{Multi-modal Translation}

 Another promising direction is multi-modal MT~\cite{yao-wan-2020-multimodal,sulubacak2020multimodal}, which involves integrating visual, audio, or other non-textual information into the translation process. This approach can enhance the quality and accuracy of translations in various settings, such as image or video captioning, automatic speech recognition, and sign language translation. LLMs, such as GPT-4~\cite{openai2023gpt4,wang2023gpt4video}, can be employed to develop models that can learn from multi-modal data and generate translations that accurately convey the meaning of the input. We demonstrate an example of multimodal translation using GPT-4-Vision~\footnote{\url{https://openai.com/research/gpt-4v-system-card}} in Figure~\ref{fig:multimodal-mt} in which the context of image must be taken into consideration when translating the sentence.

However, multi-modal MT poses several challenges, such as data heterogeneity, unbalanced datasets, and domain specificity. Overcoming these challenges would require developing novel algorithms that can learn from multi-modal data and generalize well across different modalities and domains. Leveraging the multilingual translation prowess of LLMs and combining them with models of diverse modalities unlocks the potential for remarkable applications. For instance, LLMs can be employed for video localization purposes. This tool's primary objective is to seamlessly translate video content into a desired target language while simultaneously replicating the video creator's voice using voice cloning technology for narration. Such an approach is perfectly suited for global product promotions, enabling the creation of a single video that can be effortlessly transcribed into multiple languages, catering to audiences across the world.

\begin{figure}[]
\centering
\begin{CJK}{UTF8}{gbsn}
\scalebox{0.8}{
\begin{AIbox}{GPT-4}
{\bfseries Prompt:} \\
\textit{Please translate the following text to English with emphasis on molecular biology terminology, considering the user's background and their fluency in French:} \\
\textcolor{red}{\textit{"La protéine XYZ joue un rôle crucial dans la signalisation cellulaire des plantes."}} \\

\textbf{Output}:\\
\textcolor{blue}{\textit{The XYZ protein plays a pivotal role in plant cellular signaling pathways.}}
\end{AIbox}
}
\end{CJK}
\caption{An illustration of personalized MT, adapting to user's domain-specific knowledge and language proficiency.}
\label{fig:personalized-mt-example}
\end{figure}

\section{New Evaluation Paradigm for LLMs-based MT}


Evaluating the quality of LLMs-based MT is a challenging task, as existing evaluation metrics may not be sufficient to capture the full range of translation quality~\cite{white-oconnell-1993-evaluation,isabelle-etal-2017-challenge}. In addition, existing open-access test sets may suffer from the data contamination problem as they are possibly used during the training process of LLMs~\cite{bang2023multitask_eval_chatgpt}. Evaluating on these test sets cannot correctly reflect the MT performance of LLMs. A new evaluation paradigm for LLMs-based MT should take into account the unique characteristics of LLM-based MT, such as the ability to generate fluent but inaccurate translations or the sensitivity to domain-specific knowledge. Possible approaches to a new evaluation paradigm include using specifically-designed human evaluations~\cite{graham-etal-2020-assessing,ji-etal-2022-achieving} for such systems, or even directly employ LLMs to evaluate the translation output from LLMs~\cite{kocmi2023large} - although studies show that LLMs would prefer the translation output from LLMs instead of other systems~\cite{liu2023gpteval}. Besides, using extrinsic evaluation is also feasible - the translation output can be used in other tasks and measure the corresponding performance instead of directly assessing the translation quality~\cite{moghe-etal-2023-extrinsic}. 

However, developing a new evaluation paradigm also poses several challenges. One challenge is how to balance the trade-off between evaluation efficiency and evaluation quality, as human evaluations can be time-consuming and expensive, and LLM-based evaluation can be biased. Another challenge is how to ensure the reliability and validity of the evaluation results, as different evaluators may have different subjective judgments or biases. An example of using GPT-4 to evaluate the translation output for a tweet from Elon Musk is shown in Figure~\ref{fig:overall_figure_eval}. Although GPT-4 can analyze the text based on the standards it lists, there is a certain hallucination phenomenon, which means pointing out errors that do not exist in the translation text. Overcoming these challenges requires rigorous experimental design, statistical analysis, and transparency in reporting.

\section{Privacy in MT using LLMs}

As LLMs become more powerful and widely used in MT, there are growing concerns about privacy and security~\cite{xie2023does}. In particular, LLMs may inadvertently reveal sensitive information in the source text or the translation output, such as personally identifiable information, confidential business data, or political opinions. Privacy in MT using LLMs aims to mitigate these risks by developing privacy-preserving methods that can protect the confidentiality and integrity of the translation process. One basic approach to preserve privacy in MT using LLMs is to anonymize sensitive information in the textual input and then pass it to LLMs and get the output, which is then de-anonymized. An example of such an issue using GPT-4 is shown in Figure~\ref{fig:overall_figure_privacy}. This is similar to methods integrating terminologies or user dictionaries into conventional NMT systems~\cite{Crego16systran}.

\begin{figure}[]
\centering
\scalebox{0.8}{
\begin{AIbox}{GPT-4}
{\bfseries Prompt:} \\
\textit{Translate the following Irish text to English:} \\
\textcolor{red}{\textit{"Bhí mé ag siúl sa choill nuair a chuala mé ceol álainn ón adharc."}} \\

{\bfseries Output (With LLM Integration):} \\
\textcolor{blue}{\textit{I was walking in the forest when I heard beautiful music from the horn.}} \\
{\bfseries Output (Traditional MT):} \\
\textcolor{forestgreen}{\textit{I was in the wood when I heard nice sound from the horn.}}
\end{AIbox}
}
\caption{Comparison of LLM-based MT with traditional MT for translating a nuanced Irish sentence, illustrating the finesse of LLMs in capturing intricate linguistic details.}
\label{fig:irish-mt-comparison}
\end{figure}

However, privacy-preserving methods in LLMs-based MT also pose several challenges. One challenge is how to balance the trade-off between privacy and accuracy, as privacy-preserving methods may introduce additional noise or distortion to the translation output~\cite{Dinu19training}. Another challenge is how to ensure the interoperability and compatibility of privacy-preserving methods across different languages, models, and platforms. Overcoming these challenges requires collaboration between experts in cryptography, privacy, and MT, as well as adherence to ethical and legal standards.

\section{Discussion}

\paragraph{Personalized MT}\cite{mirkin-meunier-2015-personalized,rabinovich-etal-2017-personalized} - With the advancements in LLM-based MT, the focus can be shifted towards personalized MT. This approach can enable the provision of customized translations that are tailored to each user's preferences and needs. It can include translations that are adapted to the user's language proficiency, domain-specific terminology, or cultural references. One possible approach to perform personalized MT is to prompt LLMs with user-specific preferences or metadata, such as the search histories or social media posts of the users. In other words, this aims to incorporate more contexts when translating text\cite{wang-etal-2017-exploiting-cross}. A practical illustration of this can be seen in a scenario where a user's domain expertise and language fluency guide the LLM's translation output, as demonstrated in Figure \ref{fig:personalized-mt-example}. The zero-shot ability of LLMs makes the above tasks feasible, which are difficult to achieve in previous MT systems because such data is usually unavailable and also difficult to integrate into NMT system even when it is available. However, personalized MT still raises several challenges. One of such is how to collect and store user-specific data in a privacy-preserving manner. Another critical challenge is how to measure the effectiveness of personalized MT, as traditional evaluation metrics may not capture the nuances of user preferences and needs. Overcoming these challenges requires careful consideration of ethical, legal, and technical issues.

\paragraph{Low-resource MT}~\cite{zhang2021noahnmt,haddow-etal-2022-survey,he-etal-2022-tencent} - Translation for languages with limited resources has long been a topic of interest and concern within the NLP community. The primary obstacle has been the lack of substantial parallel corpora, which is essential for training effective translation models. The advent of LLMs has brought renewed hope to this domain, given their comprehensive training on a vast array of textual data and their proven capabilities in diverse linguistic tasks. However, a deeper investigation reveals some limitations in leveraging LLMs like ChatGPT for low-resource MT. Studies, including those by \cite{bang2023multitask_eval_chatgpt}, have highlighted performance inconsistencies of LLMs for non-English languages. This is further supported by \cite{jiao2023chatgpt}, which show that the translation accuracy of LLMs suffers when addressing languages that are either low in resources or significantly divergent from English. This is largely attributed to the over-representation of English in the datasets used to train these models. Nevertheless, the extensive knowledge base and generalization abilities of LLMs present several opportunities. They can be employed in generating synthetic parallel data, thereby potentially compensating for the lack of genuine training data. An example that encapsulates this potential can be observed when translating sentences from languages like Irish, as highlighted in Figure \ref{fig:irish-mt-comparison}. Improving MT for low-resource languages is essential for preserving linguistic diversity and ensuring equitable access to information. By advancing translation capabilities for these languages, we can democratize knowledge dissemination, promote cultural understanding, and foster economic and social inclusivity on a global scale.

\section{Experiments and Analysis on Translation Performance of LLMs}
\label{sec:results}
This section presents some experimental results and analysis of various LLMs performances in translation tasks, focusing on Chinese-to-English translations. These LLMs are evaluated based on multiple criteria, including BLEU~\cite{papineni-etal-2002-bleu}, ChrF++~\cite{popovic-2017-chrf}, TER~\cite{snover2006study}, d-BLEU~\cite{liu2020multilingual} and detailed comparison across various datasets. Furthermore, an error rate analysis provides insights into specific challenges faced by LLMs.

\subsection{Chinese-to-English Translation Performance}

Table~\ref{tab:results_1} is adopted from~\newcite{jiao2023chatgpt}, showcasing the performance of different systems in Chinese-to-English translation tasks, evaluated by BLEU, ChrF++, and TER metrics. The results have demonstrated that the early version of ChatGPT~\cite{ouyangtraining-instructGPT} can achieve comparable performance compared to various specialized MT systems while it is a general-purpose dialogue system, confirming the translation capability of LLMs and the prospect of utilizing LLMs on MT.

\begin{table}[H]
\centering
\scalebox{0.95}{
\begin{tabular}{l rrr}
\toprule
\textbf{System} & \textbf{BLEU↑} & \textbf{ChrF++↑} & \textbf{TER↓} \\
\midrule
Google          & 31.66          & 57.09            & 56.21        \\
DeepL           & 31.22          & 56.74            & 57.84        \\
Tencent         & 29.69          & 56.24            & 57.16        \\
GPT-3.5         & 24.73          & 53.71            & 62.84        \\
\bottomrule
\end{tabular}
}
\caption{Comparative analysis of Chinese-to-English translation performance. The results are from~\newcite{jiao2023chatgpt}.}
\label{tab:results_1}
\end{table}

\subsection{Document-level Translation Performance}

Further analysis from~\newcite{wang2023document} on various datasets reveals how commercial MT systems and LLM applications perform on document-level Chinese-to-English translation datasets including mZPRT \cite{xu2022guofeng} and WMT2022 \cite{kocmi-etal-2022-findings} covering domains such as news, social media, web fiction, and Q\&A forum. The results shown in Table~\ref{tab:results_2} demonstrate that advanced LLMs such as GPT3.5 and GPT-4 obtained strong performance on various domains and even surpass some specialized MT systems on some domains, showcasing the competitive capability of LLMs on document-level translation.

\begin{table}[H]
\centering

\scalebox{0.96}{
\begin{tabular}{l rrrrr}
\toprule
\textbf{Model} & \textbf{News} & \textbf{Social} & \textbf{Fiction} &  \textbf{Q\&A} & \textbf{Avg} \\
\midrule
Google         & 27.7 & 35.4 & 16.0 & 12.0 & 22.8                         \\
DeepL          & 30.3& 33.4& 16.1& 11.9& 22.9                          \\
Tencent        & 29.3& 38.8& 20.7& 15.0& 26.0                      \\
GPT-3.5        & 29.1& 35.5& 17.4& 17.4& 24.9                                  \\
GPT-4          & 29.7& 34.4& 18.8& 19.0& 25.5                                   \\
\bottomrule
\end{tabular}
}
\caption{Document-level Translation performance on mZPRT \cite{xu2022guofeng} and WMT2022 \cite{kocmi-etal-2022-findings}. The results are from~\newcite{wang2023document}.}
\label{tab:results_2}
\end{table}

\subsection{Error Rate Analysis}

An in-depth error rate analysis from \newcite{wu2024adapting} on provides a clear view of the challenges LLMs face for translation, which is crucial for future development of LLMs on MT. The results in Table~\ref{tab:results_3} show that although the the finetuned LLaMA-7B~\cite{touvron2023llama1,touvron2023llama2} still exhibit various types of translation erro such as mis-translation and grammar issues, it achieves lower error rate on under-translation, omission and inconsistent style, etc. This analysis has demonstrated the potential of LLMs for translation.

\begin{table}[H]
\centering
\scalebox{0.96}{
\begin{tabular}{lrr}
\toprule
\textbf{Error Type} & \textbf{LLaMA} & \textbf{Google} \\
\midrule
Mistranslation               & 2,002             & 1,356       \\
Overtranslation              & 836               & 715         \\
Undertranslation             & 977               & 1,358       \\
Addition                     & 586               & 840         \\
Omission                     & 417               & 484         \\
Grammar                      & 474               & 266         \\
Unclear reference            & 206               & 102         \\
Cohesion                     & 933               & 704         \\
Coherence                    & 717               & 565         \\
Inconsistent style           & 85                & 771         \\
Multiple terms in translation& 386               & 339         \\
\bottomrule
\end{tabular}
}
\caption{Error rate analysis between LLaMA-7B-finetune and Google Translate. The results are from~\newcite{wu2024adapting}.}
\label{tab:results_3}
\end{table}


\section{Conclusion}

In this paper, we explored several intriguing and promising research directions for MT in the context of using LLMs. We presented discussions and case examples for long-document translation, stylised translation, interactive translation, TM-based translation, multi-modal translation and new evaluation paradigms for MT using LLMs, along with examples preserving user privacy in LLMs-based MT. Furthermore, we identified additional directions such as personalized MT . Our aim is to inspire further research in the area of leveraging LLMs for MT and to advance the state-of-the-art in this rapidly evolving field.

\section*{Acknowledgements}

We would like to thank the anonymous reviewers for their valuable suggestions and comments. Jitao Xu is supported by the China Postdoctoral Science Foundation under Grant Number 2023TQ0245. Derek F. Wong is supported by the Science and Technology Development Fund, Macau SAR (Grant Nos. FDCT/060/2022/AFJ, FDCT/0070/2022/AMJ), National Natural Science Foundation of China (Grant No. 62261160648), Ministry of Science and Technology of China (Grant No. 2022YFE0204900), and the Multi-year Research Grant from the University of Macau (Grant No. MYRG-GRG2023-00006-FST-UMDF).

\bibliographystyle{lrec_natbib}
\bibliography{lrec-coling2024-example, anthology}
\end{document}